\pdfoutput=1

\documentclass[11pt]{article}

\usepackage{acl}

\usepackage{times}
\usepackage{latexsym}

\usepackage{graphicx}
\usepackage{subcaption,booktabs}
\usepackage{tabularx}
\usepackage{multirow, amsmath}
\usepackage{placeins}
\usepackage{gensymb} 
\usepackage{amssymb}

\usepackage[T1]{fontenc}

\usepackage[utf8]{inputenc}

\usepackage{microtype}

\title{Things not Written in Text: Exploring Spatial Commonsense from Visual Signals}

\author{
	Xiao Liu$^{1}$, 
	Da Yin$^{2}$,
	Yansong Feng$^{1,3}$\thanks{\quad Corresponding author.}\and
	Dongyan Zhao$^{1,4,5}$ \\
	$^1$Wangxuan Institute of Computer Technology, Peking University\\
	$^2$Computer Science Department, University of California, Los Angeles\\
	$^3$The MOE Key Laboratory of Computational Linguistics, Peking University\\
	$^4$Artificial Intelligence Institute of Peking University\\
	$^5$State Key Laboratory of Media Convergence Production Technology and Systems\\
	{\tt \{lxlisa,fengyansong,zhaody\}@pku.edu.cn} \\
	{\tt da.yin@cs.ucla.edu}\\
}

\begin{document}
\maketitle
\begin{abstract}
Spatial commonsense, the knowledge about spatial position and relationship between objects (like \emph{the relative size of a lion and a girl}, and \emph{the position of a boy relative to a bicycle when cycling}), is an important part of commonsense knowledge. Although pretrained language models (PLMs) succeed in many NLP tasks, they are shown to be ineffective in spatial commonsense reasoning. Starting from the observation that images are more likely to exhibit spatial commonsense than texts, we explore whether models with visual signals learn more spatial commonsense than text-based PLMs. We propose a spatial commonsense benchmark that focuses on the relative scales of objects, and the positional relationship between people and objects under different actions.
We probe PLMs and models with visual signals, including vision-language pretrained models and image synthesis models, on this benchmark, and find that image synthesis models are more capable of learning \emph{accurate} and \emph{consistent} spatial knowledge than other models.
The spatial knowledge from image synthesis models also helps in natural language understanding tasks that require spatial commonsense.
Code and data are available at \url{https://github.com/xxxiaol/spatial-commonsense}.
\end{abstract}

\section{Introduction}
Spatial perception, the ability to detect the spatial position and to infer the relationship between visual stimuli~\citep{donnon2005impact, saj2015influence}, is basic but important for human beings~\citep{pellegrino1984understanding}. It is of everyday use, from understanding the surrounding environment, like \emph{when seeing a woman sitting in a car with her hands on the steering wheel, we know she is probably driving}, to processing spatial information and performing reasoning, like \emph{navigating through a dense forest}. We regard the knowledge needed in spatial perception as spatial commonsense. Humans start to develop spatial perception and acquire spatial commonsense from infancy, and apply the commonsense through lifetime~\citep{kuipers1990commonsense, poole2006development}.

Although text-based Pretrained Language Models (PLMs) achieve great performance on various commonsense reasoning tasks~\citep{davison2019commonsense, zhou2020evaluating}, they are shown to be ineffective when dealing with spatial commonsense. \citet{zhang2020language} and \citet{aroca2021prost} show that current PLMs lack the ability to reason about object scales. \citet{bhagavatula2020abductive} find that BERT~\citep{devlin2019bert} underperforms on instances involving spatial locations.
The struggle of PLMs with spatial commonsense is partly because spatial commonsense is rarely expressed explicitly in texts. We may write sentences like \emph{lions are big animals}, but we seldom explicitly mention how big lions are; we also rarely write about the spatial relationship between a boy and a bicycle when he is cycling. 

\begin{figure}[t]
\centering
\includegraphics[width=\columnwidth]{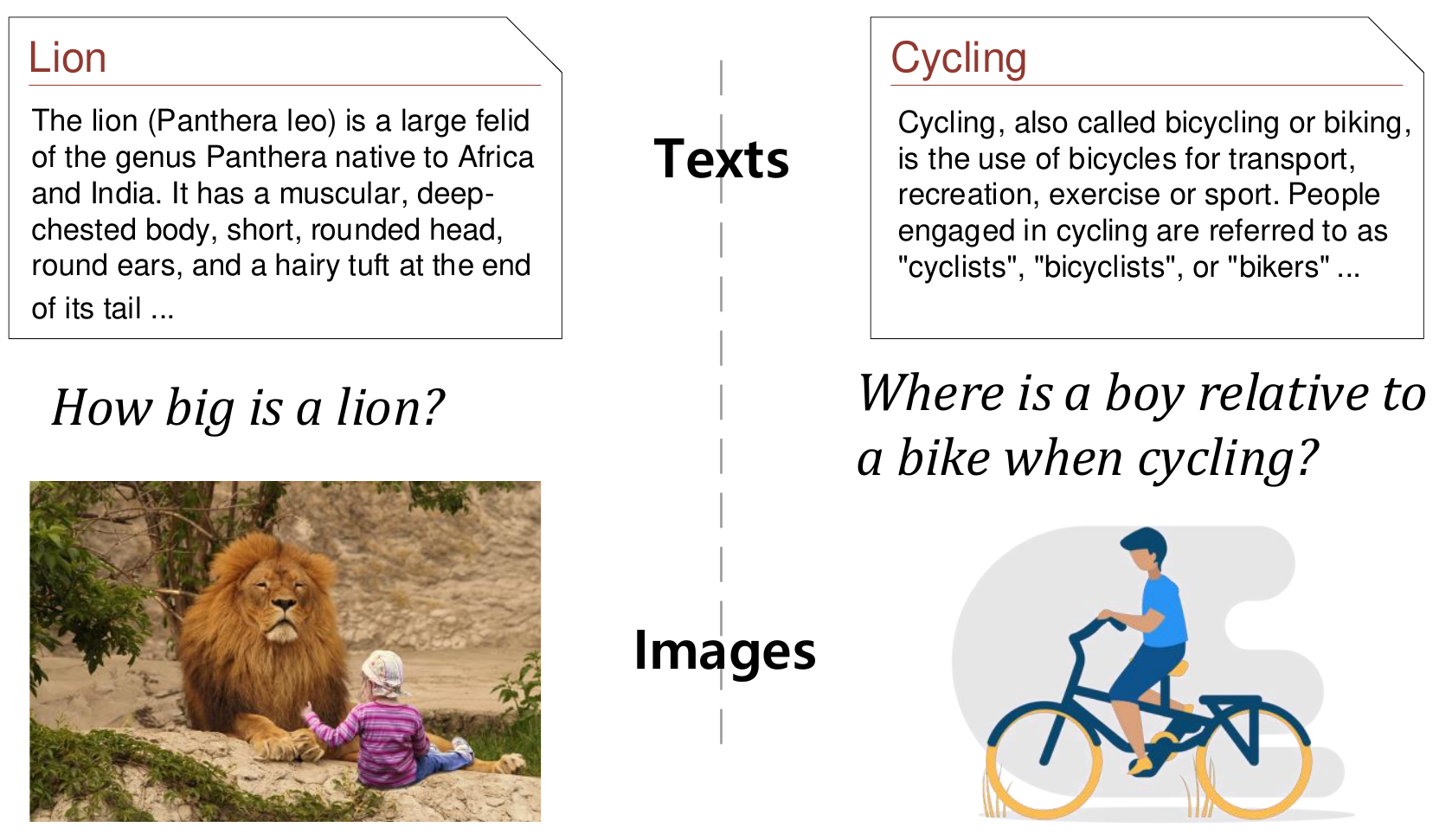} 
\caption{Texts and images related to \emph{lion} and \emph{cycling}. Images exhibit more explicit spatial knowledge than texts.}
\label{fig-intro}
\end{figure}

Spatial commonsense is exhibited in images more commonly~\citep{cui-etal-2020-beyond}.
As shown in Figure~\ref{fig-intro}, the two Wikipedia articles provide little spatial information, but a picture of \emph{a lion and a girl} provides a reference to the size of a lion; and a painting of \emph{a boy riding a bicycle} depicts that he sits \emph{on} the bicycle. Hence, a natural idea is to elicit spatial knowledge from models with visual signals.

\begin{figure*}[t]
\centering
\includegraphics[width=0.95\textwidth]{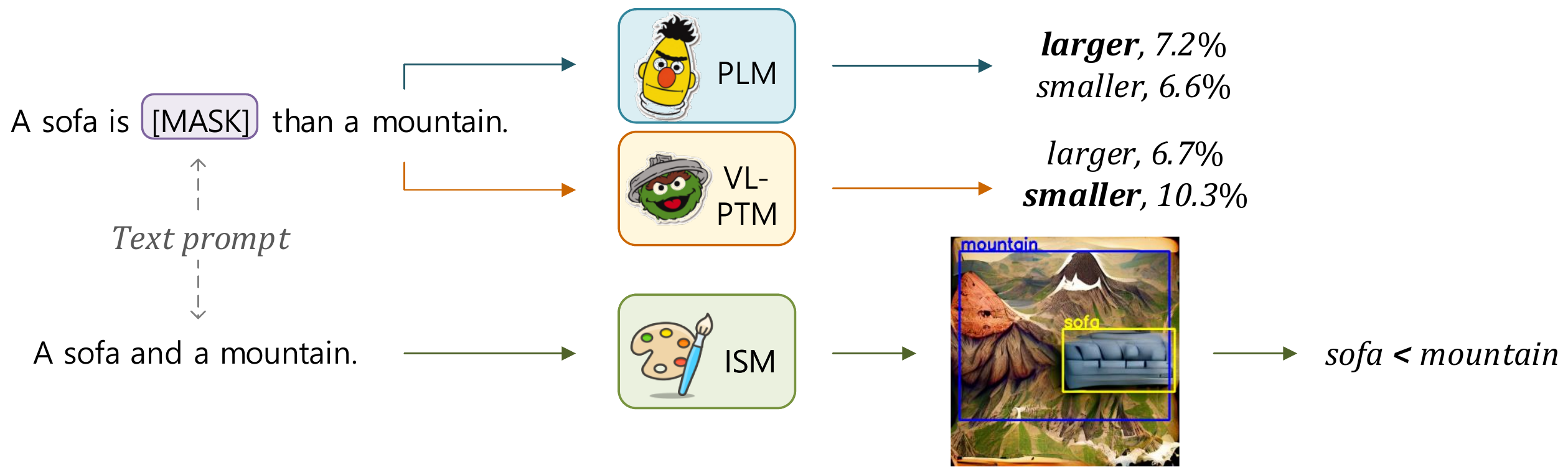} 
\caption{The probing process. We take the size comparison between $sofa$ and $mountain$ as an example.}
\label{fig-probing}
\end{figure*}
We first study \emph{whether models with visual signals learn more spatial knowledge than text-only models}. We select Vision-Language PreTrained Models (VL-PTMs) and Image Synthesis Models (ISMs) for investigation. 
VL-PTMs encode texts and images together, fusing their features to deal with downstream tasks.
ISMs take texts as input, and generate images based on the texts. 
To evaluate the spatial commonsense in PLMs and models with visual signals, we design a benchmark that involves two subtasks: 1) comparing sizes and heights of different objects (like \emph{a lion and a girl}), and 2) determining the positional relationship between a person and an object when a certain action happens (like \emph{a boy's position when riding a bicycle}). The subtasks are designed to examine the model's capability to master two kinds of spatial commonsense: understanding spatial scales, and the relationship between surrounding objects and ourselves.

As shown in Figure~\ref{fig-probing}, we probe models with text prompts on this benchmark. We feed text prompts with masks to PLMs and VL-PTMs, and take the possible word with the highest probability as their prediction.
We probe ISMs in a similar way: we first feed the text prompts to ISMs and then evaluate the generated images. We evaluate the images with two methods: automatically comparing bounding boxes of objects and conducting human evaluation. 
Results show that models with visual signals learn more accurate spatial commonsense than PLMs.

Besides the performance comparison, we are also interested in \emph{how is the quality of spatial commonsense learned by different models?} We investigate how consistent the spatial knowledge learnt by a model is, like whether it can manifest \emph{a lion is larger than a girl} and \emph{a girl is smaller than a lion} simultaneously; and to what extent models can generalize the knowledge when uncommon scenarios like \emph{an enchantress lights the sparkler} appear. 
We observe that ISMs are capable of generating consistent spatial knowledge and the performance is robust in uncommon scenarios. 

The following problem is \emph{how to benefit natural language understanding tasks with the spatial knowledge from ISMs?} We investigate this in the question answering scenario. Take a question like \emph{A boy is riding a bicycle. Is he on the bicycle?} We generate an image about the question context \emph{a boy who is riding a bicycle} with a text prompt using ISMs, and feed both the question and the generated image into vision-language models to predict an answer. This framework outperforms strong question answering models pretrained on texts only. While this is a simplified scenario of spatial commonsense reasoning, it manifests a possible way to employ the spatial knowledge learned by ISMs in natural language understanding.

Motivated by the observation that images contain more spatial commonsense than texts, we 1) design a framework, including the data and probing methods, to compare the spatial reasoning ability of models with different modalities; 2) propose methods to evaluate the quality of learned spatial commonsense, and find that models with visual signals, especially ISMs, learn more \emph{precise} and \emph{robust} spatial knowledge than PLMs; and 3) demonstrate the improvement in spatial commonsense question answering with the help of visual models.  
\section{Related Works}
\subsection{Spatial Commonsense Reasoning}
\paragraph{Object Scales.}  
\citet{bagherinezhad2016elephants} build a dataset for objects' size comparison, and \citet{elazar2019large} provide distributional information about objects' lengths. \citet{forbes2017verb} also involve spatial comparison but are criticized for ill-defined comparison~\citep{elazar2019large}. \citet{aroca2021prost} design a physical reasoning dataset that requires not only spatial commonsense but also a complex reasoning process, which is extremely challenging for existing models. We choose the formulation of object comparison in pairs as this kind of knowledge is easy to be probed from different models.

\paragraph{Spatial Relationship.}
\citet{collell2018acquiring} introduce a dataset of spatial templates for objects under different relations, but the spatial relations are represented as relative positions of bounding boxes, which are hard to express in language. \citet{yatskar-etal-2016-stating} extract statements of spatial relationship from object co-occurrences in MS-COCO~\citep{lin2014microsoft}.
\citet{mirzaee2021spartqa} design a textual spatial reasoning benchmark, and \citet{johnson2017clevr} and \citet{hudson2019gqa} involve spatial reasoning in images, but they focus on logical reasoning rather than commonsense.
Contrast to them, we build a dataset to describe the spatial relationship between people and objects in certain actions with preposition words. 
\subsection{Knowledge Probing}
Early attempts in probing PLMs~\citep{liu2019linguistic,hewitt2019structural} mainly 
train a classifier on the task of interest with the encoded representations. However, the probing performance is highly influenced by the probe design~\citep{pimentel2020information}, thus is hard to reflect the ability of PLMs.

Recently, prompt-based methods~\citep{petroni2019language, zhou2020evaluating} become more prevalent to study what knowledge PLMs already encode. PLMs take a prompt as input, and generate the continuation (for generative PLMs) or predict masked words (for discriminative PLMs). This does not need additional training, and only a small development set is used to choose optimal prompts and answers~\citep{jiang2020can}. 
In this work, we probe PLMs and VL-PTMs with prompts. 
Prompt-based methods are also used in model training~\citep{schick2021s, zhou2021learning}, while we focus on the knowledge already learned by models.

\citet{basaj2021explaining,oleszkiewicz2021visual} try to apply the probing methods into the computer vision domain, 
but they focus on probing representations of visual models. In contrast, we probe ISMs by evaluating the generated images. 
\section{Benchmark Construction}
\subsection{Datasets}
\begin{table}[t]
    \small
    \begin{subtable}[h]{0.5\textwidth}
        \centering
        \begin{tabularx}{0.85\textwidth}{l|l}
        \toprule
        \multicolumn{2}{c}{\textbf{Size}} \\
        \midrule
        1 & ant, coin, nut, bullet, dice \\
        2 & bird, cup, shell, bottle, wallet \\
        3 & tyre, chair, microwave, dog, suitcase \\
        4 & human, sofa, bookshelf, tiger, bed \\
        5 & house, cinema, mountain, truck, plane \\
        \bottomrule
       \end{tabularx}
       \caption{Objects of different levels of sizes.}
       \label{table-size}
    \end{subtable}
    \begin{subtable}[h]{0.5\textwidth}
        \centering
        \begin{tabularx}{0.85\textwidth}{l|l}
        \toprule
        \multicolumn{2}{c}{\textbf{Height}} \\
        \midrule
        1 & ant, insect, water drop, bullet, dice \\
        2 & bird, cup, shoe, bottle, mobile phone \\
        3 & table, chair, trash can, sofa, suitcase \\
        4 & human, horse, bookshelf, camel, door \\
        5 & apartment, theatre, giraffe, truck, street lamp \\
        \bottomrule
       \end{tabularx}
       \caption{Objects of different levels of heights.}
       \label{table-height}
    \end{subtable}
    \caption{The dataset of object scales.}
    \label{table-dimension}
\end{table}
\paragraph{Size and Height.}
Inspired by the cognitive discovery~\citep{hersh1976fuzzy} that people tend to categorize objects scales into fuzzy sets, we select 25 common objects in daily life, and categorize them into 5 groups as shown in Table~\ref{table-size} to construct the dataset for size comparison. Typical objects in the former group are smaller than those in the latter group.
We form 250 pairs of objects from different groups, like $\langle$\emph{ant, bird}$\rangle$, where the first object is smaller than the second in commonsense. Models are asked to compare the size of objects in pairs. To avoid an imbalance of answer distribution, we also consider the reversed pairs like $\langle$\emph{bird, ant}$\rangle$, so there are 500 instances in total.

The dataset for comparing objects' heights is constructed similarly, as shown in Table~\ref{table-height}. We also form 500 instances with the objects. The comparison between objects is validated by 5 human annotators for both datasets.

\begin{figure}[t]
\centering
\includegraphics[width=0.95\columnwidth]{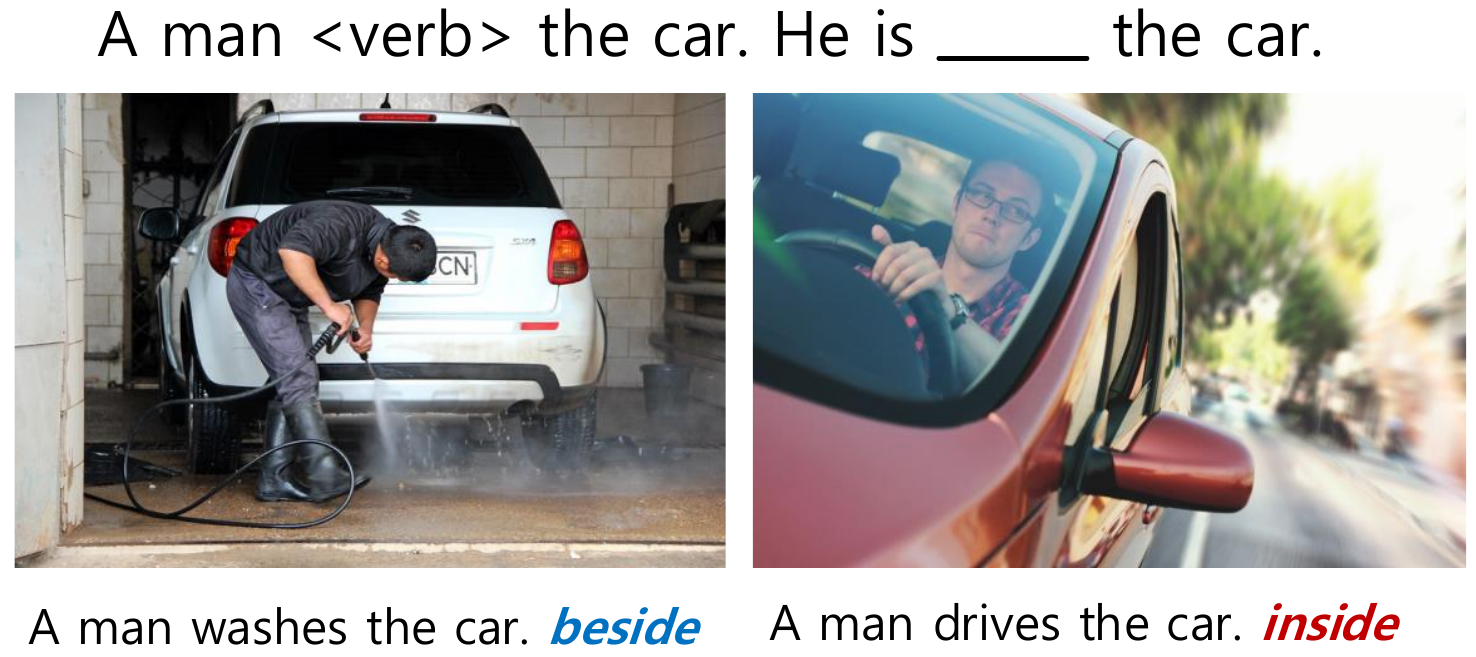} 
\caption{Example of two positional relations between \emph{man} and \emph{car}.}
\label{fig-position}
\end{figure}
\paragraph{Positional Relationship.}
The positional relationship dataset consists of human actions regarding objects and the most likely positional relation between the person and the object. We consider four types of positional relations: \emph{above, below, inside, beside}, as they do not overlap with each other. 

We select common objects, and write actions between people and the objects. The actions do \emph{not} contain prepositions, like \emph{sit on the chair}. Each object is accompanied by two actions with different positional relations. Take Figure~\ref{fig-position} as an example. The man is \emph{beside} the car when washing the car, whereas he is \emph{inside} the car when driving it. Therefore, the relation cannot be easily inferred from collocations between the person and the object.
The dataset contains 224 instances, validated by 5 annotators.
\begin{table*}[t!]
\centering
\small
\begin{subtable}[h]{0.45\textwidth}
    \centering
    \begin{tabular}{lcc}
    \toprule
    \textbf{Model} & \textbf{Acc} (avg. / $\sigma$) & \textbf{F1} (avg. / $\sigma$)\\
    \midrule
    BERT & 49.8 / 2.66 & 47.7 / 2.48\\
    RoBERTa & 54.1 / 3.93 & 52.2 / 6.92\\
    VinVL & \textbf{61.8} / 2.47 & \textbf{54.4} / 3.06\\
    \midrule
    \textbf{Model} & \textbf{Acc} & \textbf{F1}\\
    \midrule
    Best PLM$^{\S}$ & 54.1 (52.2) & 52.2 (46.7) \\
    VinVL$^{\S}$ & \textbf{61.8} (61.6) & 54.4 (53.8) \\
    ISM (Box)$^{\S}$ & 54.8 (\textbf{81.6}) & \textbf{54.8} (\textbf{81.6}) \\
    \midrule
    Best PLM$^{\dag}$ & 54.1 (52.9) & 52.2 (51.0) \\
    VinVL$^{\dag}$ & 61.8 (61.6) & 54.4 (54.3) \\
    ISM (Human)$^{\dag}$ & \textbf{72.7} (\textbf{76.5}) & \textbf{72.6} (\textbf{76.4}) \\
    \bottomrule
    \end{tabular}
    \caption{Comparing sizes of objects. Both objects are recognized by the object detection model in 15\% images and are recognized by humans in 86\% images.}
\end{subtable}
\hspace{1em}
\begin{subtable}[h]{0.45\textwidth}
    \centering
    \begin{tabular}{lcc}
    \toprule
    \textbf{Model} & \textbf{Acc} (avg. / $\sigma$) & \textbf{F1} (avg. / $\sigma$)\\
    \midrule
    BERT & 50.8 / 2.29 & 50.3 / 0.25\\
    RoBERTa & 50.8 / 6.43 & 49.2 / 7.45\\
    VinVL & \textbf{64.5} / 7.61 & \textbf{61.5} / 10.5\\
    \midrule
    \textbf{Model} & \textbf{Acc} & \textbf{F1}\\
    \midrule
    Best PLM$^{\S}$ & 50.8 (48.6) & 50.3 (47.9) \\
    VinVL$^{\S}$ & \textbf{64.5} (\textbf{69.3}) & \textbf{61.5} (65.2)\\
    ISM (Box)$^{\S}$ & 52.5 (68.1) & 52.5 (\textbf{68.1}) \\
    \midrule
    Best PLM$^{\dag}$ & 50.8 (48.5) & 50.3 (47.5) \\
    VinVL$^{\dag}$ & 64.5 (63.9) & 61.5 (60.6)\\
    ISM (Human)$^{\dag}$ & \textbf{78.9} (\textbf{85.4}) & \textbf{78.8} (\textbf{85.3}) \\
    \bottomrule
    \end{tabular}
    \caption{Comparing heights of objects. Both objects are recognized by the object detection model in 14\% images and are recognized by humans in 82\% images.}
\end{subtable}
\caption{Probing performance on object scales. The numbers are in percentages (\%). The number before the slash (/) is the average performance of different folds, and the number after the slash is the standard deviation. The number out of parentheses is the performance on the whole dataset, and the number in parentheses indicates performance on the subset of instances where the generated images can be recognized by object detection models ($^{\S}$), and on the subset recognized by humans ($^{\dag}$). }
\label{table-probing-size}
\end{table*}
\subsection{Probing Tasks}
We probe PLMs and VL-PTMs through masked word prediction. Given a text prompt with masks and a set of possible words, a model calculates the probability of each possible word filling the masked position. The word with the highest possibility is regarded as the prediction.

We also probe ISMs through text prompts. The input is a piece of descriptive text, and the output is the image generated by an ISM. We assess the image with two methods as described in~\ref{subsection-evaluation}.

PLMs are found to perform poorly in scenarios involving complex reasoning over spatial knowledge~\citep{aroca2021prost}, and we want to investigate whether they even fail in early stages, like whether they have learned spatial knowledge. So we probe models with simple tasks.
In the subtask of size and height, the prompt for PLMs and VL-PTMs is in the form of \emph{$O_a$ is [MASK] than $O_b$}, where $\langle O_a, O_b \rangle$ is an object pair. The possible answer set is $\{larger, smaller\}$ for size and $\{taller, shorter\}$ for height. The prompt for ISMs is in the form of \emph{$O_a$ and $O_b$}, and the objects in generated images are compared for size and height.

In the subtask of positional relationship, the prompt for PLMs and VL-PTMs contains an event scenario and a masked token for the positional relationship, like \emph{A woman washes the car. She is [MASK] the car.} The possible answer set is $\{above, below, inside, beside\}$. The prompt for ISMs describes the scenario only, like \emph{A woman washes the car.} 

\subsection{ISM Evaluation}
\label{subsection-evaluation}
We assess the images generated by ISMs with two methods.
We first use the spatial information of bounding boxes (referred to as ISM (Box)). For each object mentioned in the prompt, we select the classified bounding box with the highest confidence. To mitigate the effect of viewpoint (an object closer to the camera may appear larger in the image), we compute the average depth of the box as the object's depth. We use the object detector from \citet{zhang2021vinvl}, and the depth estimator from \citet{godard2019digging}.
When probing the relative size, we compare $area \times depth^2$ of the two objects' boxes; and when probing the relative height, we compare $height \times depth$. When classifying positional relations, we use the mapping rules between spatial relations and image regions from Visual Dependency Grammar (VDG)~\citep{elliott2013image}. We list the rules in Appendix~\ref{appendix-vdg}.

Some generated images are vague while object detection models are trained to process clear pictures, so a number of objects are not recognized. To precisely assess the generated images, we conduct human evaluation on all images (referred to as ISM (Human)). Annotators are asked to compare the size/height of the objects in the images (for the first subtask) and classify the positional relationship between the person and the object (for the second subtask). Each image is evaluated by two annotators, and the average performance is reported. Specifically, we report the accuracy and macro F1 between models' predictions and correct answers. Besides the performance of ISMs on the subset of recognized instances, we also report the performance on the full dataset, giving the unrecognized instances a random guess.
\section{Probing Spatial Commonsense}
\subsection{Models}
We take BERT~\citep{devlin2019bert} and RoBERTa~\citep{liu2019roberta} as examples of text-only PLMs. 
For VL-PTMs, we choose VinVL~\citep{zhang2021vinvl}, which performs well in various vision-language tasks. It uses a transformer-based backbone and is pretrained on various vision-language datasets including image caption datasets, visual QA datasets, etc. As it preserves the masked word prediction objective like PLMs, it can also be probed with prompts.
We choose VQGAN+CLIP\footnote{Developed simultaneously by Ryan Murdoch and~\citet{crowson2022vqgan}. Implementation details are in Appendix~\ref{appendix-image}.} as a representative of ISMs. It uses CLIP~\citep{radford2021learning} to guide VQGAN~\citep{esser2021taming} to generate images that best match the given text. 
To make a fair comparison regarding model size, we select BERT-large, RoBERTa-large, and VinVL-large. We use VQGAN with codebook size $Z=16384$ and downsampling factor $f=16$, and CLIP with ViT-B/32~\citep{dosovitskiy2020image} architecture. All four models are of similar sizes. 

As language models are sensitive to the expressions in probing~\citep{liu2021pre} (like changing an answer choice from \emph{larger} to \emph{bigger}, the predictions of BERT may differ a lot), we generate new prompts and answers based on those originally designed in the benchmark, and search for the optimal ones for PLMs and VL-PTMs. Similar to \citet{jiang2020can}, we use back-translation to generate 10 candidates for prompts and answers, and filter out the repeated ones. To select prompts and answers, we split the dataset into 5 folds, where different folds do not share the same objects. For each run, one fold is used as the development set to choose the best candidate, and the model is probed on other folds with the chosen prompt. We report average performance of 5 runs.
\subsection{Probing Results}
\begin{figure}[t]
\centering
\includegraphics[width=0.95\columnwidth]{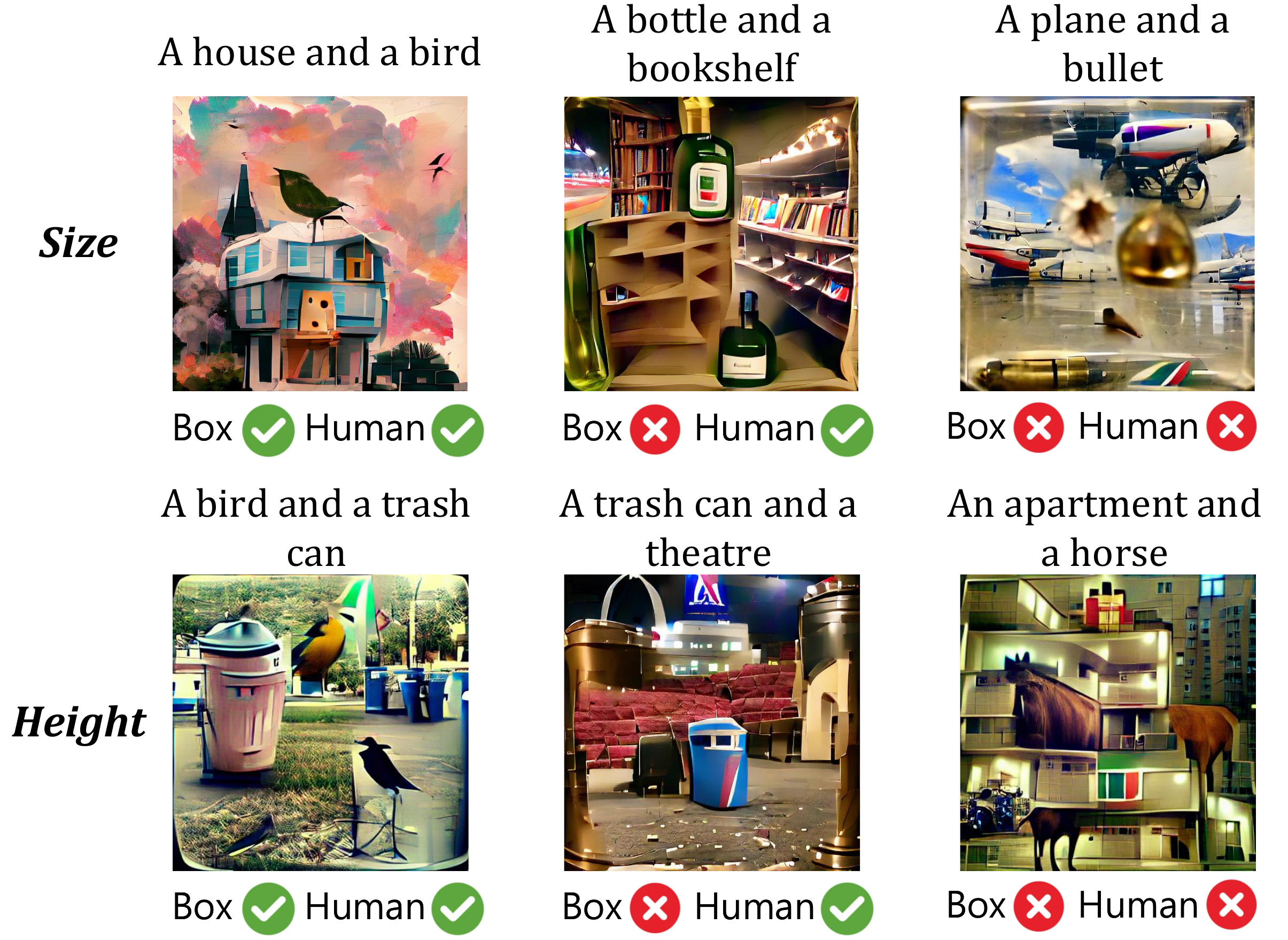} 
\caption{Images generated by ISM in scale comparison. \checkmark means objects are successfully recognized by the object detection model or humans, and $\times$ means not.}
\label{fig-size-image}
\end{figure}
\paragraph{Size and Height.}
Table~\ref{table-probing-size} reports the probing performance of comparing the scales of objects. We also demonstrate probing results on RelativeSize~\citep{bagherinezhad2016elephants} in Appendix~\ref{appendix-relativesize}.
We observe that PLMs perform similarly. Even the best PLMs are slightly better than random guesses, indicating they are ineffective in predicting object scales.
Although RoBERTa is trained on more texts and assumed to encode more knowledge, its performance is similar to BERT's. It shows that PLMs do not learn much spatial commonsense from texts even if the pretrained corpus greatly increases.

With the help of visual features in pretraining, VinVL greatly outperforms PLMs. ISM (Box), which simply compares bounding boxes in images generated by the ISM, also outperforms PLMs. Since only a small portion of instances are recognized with bounding boxes, if we only consider the predictions on these instances, the gap between ISM (Box) and PLMs is more than 15\%. These indicate that models with visual signals learn accurate spatial commonsense knowledge from images. 

ISM (Box) outperforms VinVL on those recognizable instances (81.6 vs. 53.8), but the recognition ratio is admittedly low. We conduct human evaluation on the generated images for more precise assessment. More than 80\% of images are recognized by humans and these images accurately reflect the spatial commonsense compared to PLMs and VinVL.~\footnote{The agreement between annotators is more than 90\%.}
The gap between VinVL and ISM (Human) may be due to different ways of using visual signals in pretraining. A training objective of VinVL, and other VL-PTMs, is aligning text with image regions. The discriminative features of objects are amplified, while other features may not receive as much attention. For instance, the shape and color are the discriminative features of an \emph{apple}, and its size is not that important in recognition. In image synthesis, models need to learn comprehensive knowledge of objects for reconstruction, and spatial knowledge may be learned implicitly in this process. 

Figure~\ref{fig-size-image} demonstrates images generated by the ISM given the prompts of object pairs. ISM grasps the main characteristics of the objects, including their scales. Some objects (like \emph{theatre} at the bottom of the middle column) can be identified by humans but are difficult for the object detection model because they are obstructed by objects in the foreground. And some objects are generated in multiple fragments (like \emph{plane} and \emph{horse} in the right column), therefore cannot be recognized by either the object detection model or humans.
\begin{table}[t!]
    \centering
    \small
    \begin{tabular}{lcc}
    \toprule
    \textbf{Model} & \textbf{Acc} (avg. / $\sigma$) & \textbf{F1} (avg. / $\sigma$)\\
    \midrule
    BERT & 26.1 / 4.15 & 19.0 / 5.20\\
    RoBERTa & 31.0 / 15.4 & 20.1 / 9.29 \\
    VinVL & \textbf{56.1} / 7.09 & \textbf{41.8} / 6.69 \\
    \midrule
    \textbf{Model} & \textbf{Acc} & \textbf{F1}\\
    \midrule
    Best PLM$^{\S}$ & 31.0 (32.5) & 20.1 (17.6) \\
    VinVL$^{\S}$ & \textbf{56.1} (\textbf{56.0}) & \textbf{41.8} (\textbf{36.0}) \\
    ISM (Box)$^{\S}$ & 33.0 (42.5) & 26.5 (26.1) \\
    \midrule
    Best PLM$^{\dag}$ & 31.0 (30.5) & 20.1 (20.1) \\
    VinVL$^{\dag}$ & 56.1 (56.4) & 41.8 (42.9) \\
    ISM (Human)$^{\dag}$ & \textbf{73.4} (\textbf{75.4}) & \textbf{65.1} (\textbf{68.0}) \\
    \bottomrule
    \end{tabular}
    \caption{Probing performance on positional relationship (\%). The symbols are identical to those in Table~\ref{table-probing-size}. Both the person and the object are recognized with bounding boxes in 39\% images and by humans in 93\% images.}
    \label{table-probing-position}
\end{table}
\paragraph{Positional Relationship.}
The probing performance on positional relationship is shown in Table~\ref{table-probing-position}. VinVL outperforms PLMs more than 20\%, and ISM (Human) outperforms PLMs more than 35\%, suggesting that models with visual signals learn more knowledge of the scenarios, especially the positions of objects relative to people.

The gap between PLMs and ISM (Box) is smaller compared to the gap in the subtask of size and height. One reason is that the rules defined in VDG cannot perfectly reflect the true positional relationship in images. For example, the man is \emph{beside} the car in the left image of Figure~\ref{fig-position}, but he will be regarded as \emph{inside} the car by the rules, as the region of car covers the region of man. 

Text-based PLMs tend to lean towards certain positional relations between a person and an object, without referring to the action. In 64\% cases, RoBERTa chooses the same option for a $\langle$\emph{person, object}$\rangle$ pair with different actions, while the proportion is 21\% for VinVL, and 28\% for ISM (Human).
\vspace{0.3cm}
\section{Quality of Spatial Knowledge}
\subsection{Consistency}
\begin{figure*}[htb]
\begin{subfigure}{\textwidth}
\includegraphics[width=0.95\textwidth]{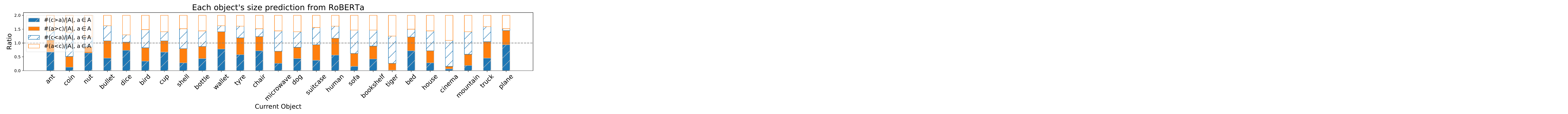} 
\end{subfigure}
\begin{subfigure}{\textwidth}
\includegraphics[width=0.95\textwidth]{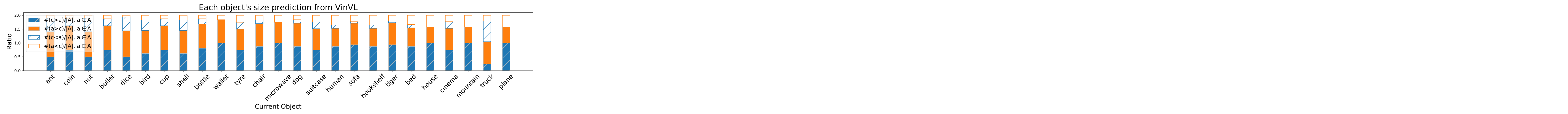} 
\end{subfigure}
\caption{Predictions from RoBERTa and VinVL in the subtask of objects' sizes. $c$ is the current object and $A$ is the set of all other comparable objects. $\#(c>a)/|A|$ indicates the ratio of predicting the current object larger than others. As $c>a$ and $a>c$ should not appear simultaneously, the sum of the two solid bars is expected to be 1.}
\label{fig-sym}
\end{figure*}
\begin{table}[t]
    \centering
    \small
    \begin{tabular}{lcccc}
    \toprule
    \multirow{2}{*}{\textbf{Model}} & \multicolumn{2}{c}{Size} & \multicolumn{2}{c}{Height} \\
    & Sym. & Trans. & Sym. & Trans. \\
    \midrule
    Best PLM & 37.5 & 71.9 & 25.9 & 73.1 \\
    VinVL & \textbf{43.5} & \textbf{95.0} & \textbf{43.0} & \textbf{93.2} \\
    \midrule
    Best PLM$^{\dag}$ & 36.6 & 72.2 & 26.1 & 72.3 \\
    VinVL$^{\dag}$ & 44.4 & \textbf{95.3} & 32.2 & \textbf{97.8} \\
    ISM (Human)$^{\dag}$ & \textbf{82.5} & 81.1 & \textbf{83.2} & 85.2 \\
    \bottomrule
    \end{tabular}
    \caption{The percentage (\%) of predictions that meet consistency. Sym and Trans indicate symmetry and transitivity. $^{\dag}$ indicates performance on the subset of images recognized by humans.}
    \label{table-consistency}
\end{table}

Models that master better spatial knowledge should be able to infer the relative scale of two objects from intermediate references. 
For example, if a model knows \emph{a dog is larger than an ant} and \emph{a sofa is larger than a dog}, it may learn \emph{a sofa is larger than an ant}, even if it has not seen \emph{sofa} and \emph{ant} together. We inspect models on how consistent their probing results are.

The consistency is measured in two aspects: \emph{symmetry} and \emph{transitivity}. Symmetry implies that if a model predicts $A>B$, then it should also predict $B<A$, and vice versa: $A<B \implies B>A$. Here $>$ and $<$ are in terms of size or height. We enumerate the object pairs and count the percentage of predictions that meet the symmetry criterion. Transitivity implies that if a model predicts $A>B$ and $B>C$, then it should predict $A>C$. It also works for $<$, $A<B \ \land\ B<C \implies A<C$. We enumerate the triples $\langle A, B, C\rangle$ where the predicted relation between $\langle A, B \rangle$ is identical to the prediction between $\langle B, C \rangle$, and count the percentage that the prediction between $\langle A, C \rangle$ meets the transitivity criterion. Note that we only evaluate whether the predictions are consistent with each other, regardless of the gold answers. 

We evaluate the consistency of predictions from PLMs that perform the best in the probing tasks (RoBERTa for size and BERT for height), VinVL, and ISM (Human). The results are in Table~\ref{table-consistency}.

VinVL outperforms the best PLM in both metrics, and the characteristics of them are similar: the transitive consistency is high, while the symmetric consistency is low. To further analyze this phenomenon, we exhibit each object's size predictions from RoBERTa and VinVL in Figure~\ref{fig-sym}. The models exhibit different behaviors in recognizing object scales. 
As the objects (X-axis of Figure~\ref{fig-sym}) are roughly listed from smaller to larger groups, the bottom blue bars are expected to follow a non-descending order from left to right, and the solid orange bars should be non-ascending. The predictions of VinVL are generally in line with this trend, while RoBERTa's predictions are disordered. For example, \emph{ant} is predicted to be \emph{larger than} other objects with high probability, and \emph{cinema is larger than others} is unlikely to happen.
On the other hand, if the model predictions are consistent, the two solid bars should sum to 1. However, the sum is far above 1 for most objects in VinVL's predictions. This bias towards words indicating the choice of \emph{large} may come from the pretraining corpus. For example, \emph{sofa} occurs twice as many times with words indicating large as with words indicating small in COCO~\citep{lin2014microsoft}, one of VinVL's pretraining datasets.

\begin{figure*}[htb]
\centering
\begin{subfigure}{0.72\columnwidth}
\centering
\includegraphics[width=\textwidth]{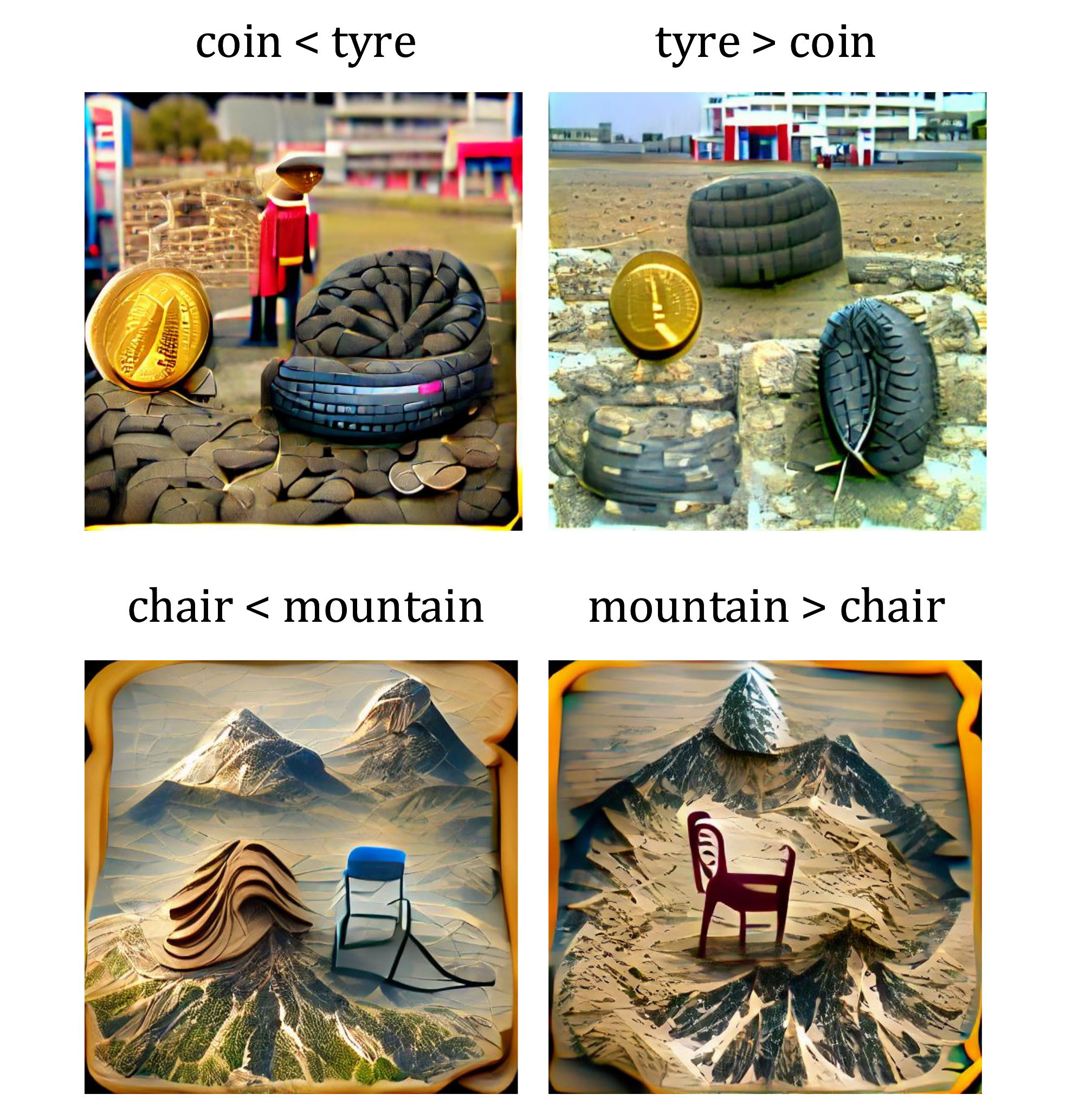} 
\caption{Two groups of generated images. Sizes of objects are consistent with each other.}
\label{fig-ism-sym}
\end{subfigure}
\hspace{0.7em}
\begin{subfigure}{1.08\columnwidth}
\centering
\includegraphics[width=\textwidth]{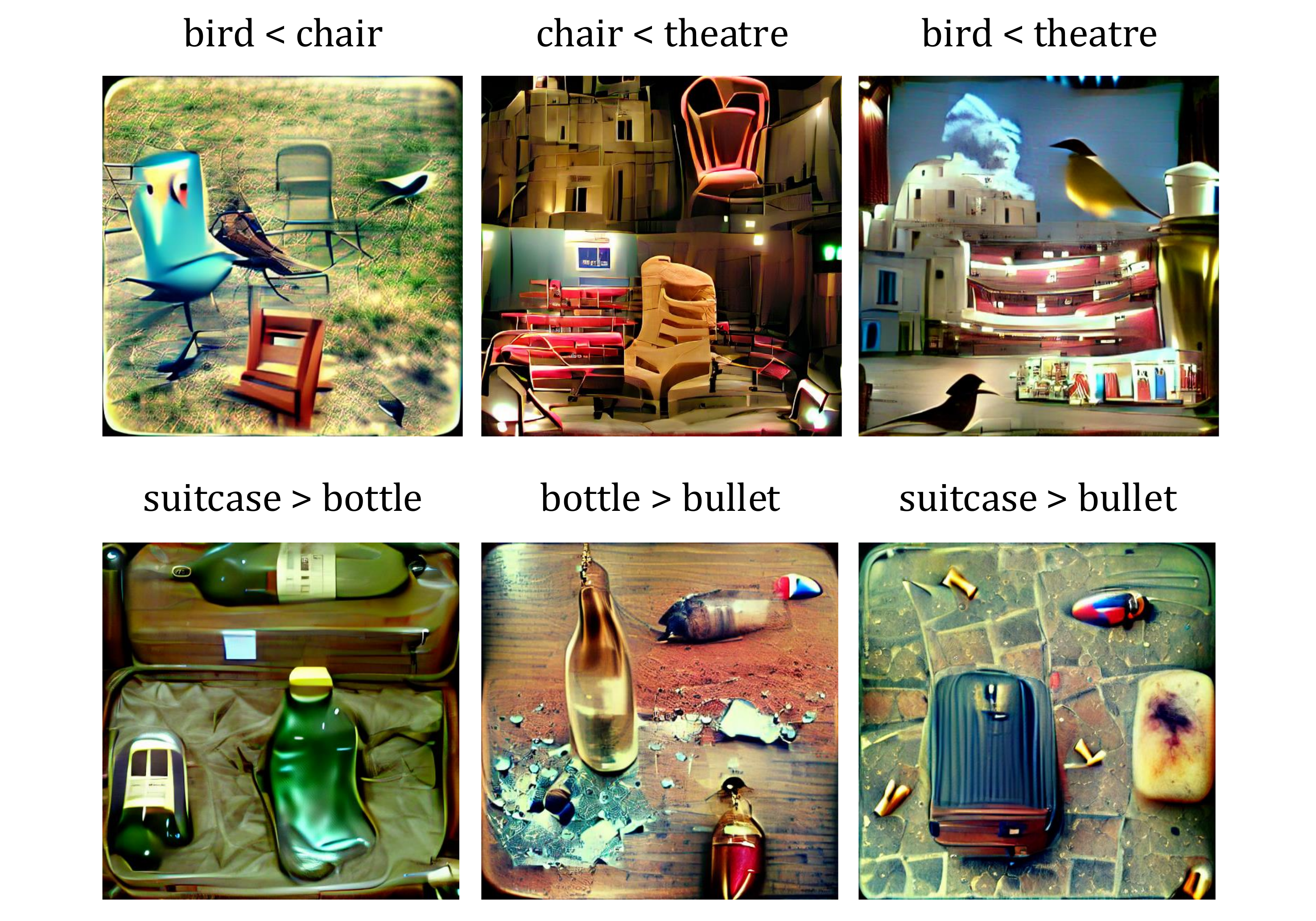}
\caption{Two groups of generated images. Heights of objects meet the transitivity criterion.}
\label{fig-ism-trans}
\end{subfigure}
\caption{Examples of the symmetric and transitive consistency of images generated by ISM.}
\label{fig-ism-cons}
\end{figure*}

ISM's predictions comply with the symmetry criterion, outperforming other models by 40\%, while also having good transitive consistency. The knowledge probed from ISM is more consistent. 
Figure~\ref{fig-ism-cons} exhibits the symmetric and transitive consistency of images generated by ISM. 
The consistency of scale knowledge makes the predictions more convincing, and gives models a chance to learn new comparisons between objects.



\subsection{Generalizability}
\begin{table}[t]
    \centering
    \small
    \begin{tabular}{lcc}
    \toprule
    \textbf{Model} & \textbf{Acc} (avg. / $\sigma$) & \textbf{F1} (avg. / $\sigma$)\\
    \midrule
    BERT & 27.4 / 3.17 & 19.7 / 7.25 \\
    RoBERTa & 29.5 / 16.0 & 20.1 / 9.90 \\
    VinVL & \textbf{58.1} / 1.97 & \textbf{44.4} / 1.63 \\
    \midrule
    \textbf{Model} & \textbf{Acc} & \textbf{F1} \\
    \midrule
    Best PLM$^\dag$ & 29.5 (28.4) & 20.1 (19.1) \\
    VinVL$^\dag$ & 58.1 (52.3) & 44.4 (41.0) \\
    ISM (Human)$^\dag$ & \textbf{66.5} (\textbf{74.8}) & \textbf{59.4} (\textbf{69.2}) \\
    \bottomrule
    \end{tabular}
    \caption{Probing models on the generalized dataset of positional relationship. The symbols are identical to those in Table~\ref{table-probing-size}. The human recognition ratio is 81\%.}
    \label{table-generalization}
\end{table} 
ISM may learn positional relations from training images directly. For example, \emph{a boy riding a bicycle} is a \emph{common} scenario and may frequently exist in ISM's training set, so models can generate images more easily when being fed with the text prompts like \emph{a boy rides a bicycle}. To further challenge ISM's capability, we make a generalized version of our original positional relationship dataset. It is designed to examine whether models are able to 
robustly reflect the spatial commonsense knowledge when facing \emph{uncommon} scenarios.

A generalized scenario is built upon the original one by replacing the person and object in the text prompts. We select the new person and new object from the subterms of the original ones (those with \emph{IsA} relation in ConceptNet~\citep{speer2017conceptnet}, like \emph{enchantress} is a \emph{woman}). To ensure these newly constructed scenarios are not likely to appear in the training data of models, we search them in BookCorpus~\citep{zhu2015aligning} and remove the scenarios that have appeared. The newly generated scenarios are also validated by humans to ensure that they are reasonable.

\begin{figure}[t]
\centering
\includegraphics[width=0.95\columnwidth]{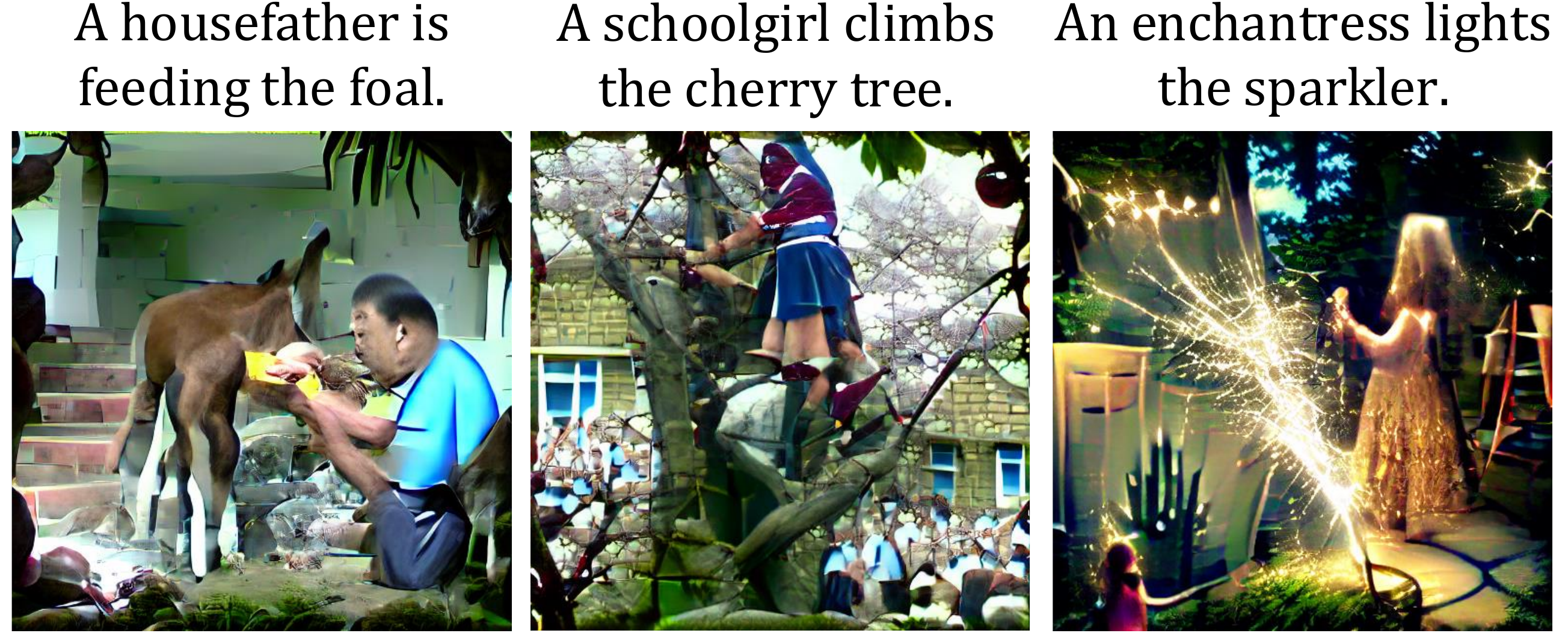} 
\caption{Images generated by ISM with the generalized prompts.}
\label{fig-generalize}
\end{figure}

Results of probing PLMs, VinVL, and ISM\footnote{We do not consider ISM (Box) because many new objects we used are unfamiliar to object detection models. Only 17\% of the objects are in the object detection classes.} on the generalized dataset are in Table~\ref{table-generalization}. 
PLMs and VinVL achieve similar performance on both the generalized dataset and the original one, indicating that they behave robustly when facing uncommon scenarios.
The performance gap between other models and ISM (Human) slightly narrows down, but ISM (Human) still outperforms VinVL more than 8\%.
Figure~\ref{fig-generalize} exhibits images generated by ISM with the generalized prompts. 
Although it is difficult for ISM to generate unfamiliar objects, it is still capable of capturing the positional relations.

\section{Solving Natural Language Questions}
We investigate how to acquire spatial knowledge from ISMs and whether the knowledge is effective in natural language understanding scenarios.
To our best knowledge, there is no appropriate task that focuses on spatial commonsense, so we create a toy task by transforming our probing benchmark into the form of question answering (QA). 
\paragraph{Dataset.}
We construct a QA dataset of yes/no questions.
Questions of objects' sizes are in the form of \emph{Is $O_a$ larger/smaller than $O_b$?} And questions of objects' heights are like \emph{Is $O_a$ taller/shorter than $O_b$?}, where $O_a$ and $O_b$ are two objects. Questions about positional relationship are accompanied with the action: for instance, \emph{A man washes the car. Is the man inside the car?} To avoid bias in answer distribution, the numbers of \emph{yes} and \emph{no} are equal in gold answers. There are 500 questions for size, 500 for height, and 448 for positional relationship.

\paragraph{Models.}
We use VinVL-base together with our image synthesis model VQGAN+CLIP to answer spatial commonsense questions. The VinVL here is finetuned on the VQA~\citep{goyal2017making} task. It takes images generated from ISM with textual prompts from questions, and predicts the answer based on the question and image together. Note that the VQA training corpus does not contain commonsense reasoning questions.

We choose UnifiedQA~\citep{khashabi2020unifiedqa} as a text-based QA model for comparison. Based on the pretrained T5 model~\citep{raffel2019exploring}, UnifiedQA is continually trained on various QA tasks, including three yes/no datasets.
We use UnifiedQA-large, which is comparable with our synthesis and reasoning model (ISM w/ VinVL) in size.
\begin{table}[t]
    \centering
    \small
    \begin{tabular}{p{2cm}p{0.4cm}p{0.4cm}p{0.4cm}p{0.4cm}p{0.4cm}p{0.4cm}}
    \toprule
    \multirow{2}{*}{\textbf{Model}} & \multicolumn{2}{c}{Size} & \multicolumn{2}{c}{Height} & \multicolumn{2}{c}{PosRel}\\
    & Acc & F1 & Acc & F1 & Acc & F1 \\
    \midrule
    UnifiedQA & 51.3 & 38.5 & 58.4 & 52.8 & 56.7 & 48.1\\
    ISM w/ VinVL & \textbf{52.4} & \textbf{43.8} & \textbf{59.4} & \textbf{54.3} & \textbf{59.8} & \textbf{58.7}\\
    \bottomrule
    \end{tabular}
    \caption{Performance of answering commonsense questions. Accuracy (\%) and macro F1 (\%) are reported. PosRel refers to positional relationship.}
    \label{table-qa}
\end{table}
\paragraph{Results.}
As shown in Table~\ref{table-qa}, ISM w/ VinVL outperforms UnifiedQA on all subtasks, showing that spatial knowledge from ISMs can be directly used by vision-language models without additional training.
Although some images cannot be precisely recognized by object detection models, vision-language models may find regions that are related to the objects mentioned in questions, and make decisions based on the features of these regions.
The results on the simple natural language task show that it is beneficial to tackle natural language tasks with vision-language methods, and ISMs can be \emph{a bridge between the two modalities}.
With the development of ISMs and object detection techniques, we believe the generated images will help more.
\section{Conclusion}
We propose a new spatial commonsense probing framework to investigate object scales and positional relationship knowledge in text-based pretrained models and models with visual signals. Experimental results show that models with visual signals, especially ISMs, learn more accurate and consistent spatial commonsense than text-only models. Integrating ISMs with visual reasoning models outperforms PLMs in answering spatial questions. This manifests the potential of using spatial knowledge from ISMs in natural language understanding tasks. 

\section*{Acknowledgments}
This work is supported in part by National Key R\&D Program of China (No. 2020AAA0106600) and NSFC (62161160339). We would like to thank the anonymous reviewers and action editor for the helpful discussions and suggestions. Also, we would thank Quzhe Huang, Chen Zhang, Chen Henry Wu, Yuxuan Lai and Nan Hu for their detailed comments. For any correspondence, please contact Yansong Feng.

\bibliography{anthology}
\bibliographystyle{acl_natbib}
\clearpage

\appendix
\section{Implementation Details}
\label{appendix-implementation}
\begin{table}[htp]
    \centering
    \small
    \begin{tabular}{lp{5.3cm}}
    \toprule
    \textbf{Relation} & \textbf{Definition} \\
    \midrule
    X \emph{inside} Y & The entirety of region X overlaps with Y. \\
    X \emph{beside} Y & The angle between the centroid of X and the centroid of Y lies between 315$\degree$ and 45$\degree$ or 135$\degree$ and 225$\degree$. \\
    X \emph{above} Y & The angle between X and Y lies between 225$\degree$ and 315$\degree$. \\
    X \emph{below} Y & The angle between X and Y lies between 45$\degree$ and 135$\degree$. \\
    \bottomrule
    \end{tabular}
    \caption{Spatial relations between image regions in Visual Dependency Grammar (VDG).}
    \label{table-vdg}
\end{table}
\subsection{Spatial Relations in Visual Dependency Grammar} 
\label{appendix-vdg}
We use the rules defined in Visual Dependency Grammar~\citep{elliott2013image} to determine the positional relationship between bounding boxes. The rules used are listed in Table~\ref{table-vdg}. If two bounding boxes meet the \emph{inside} standard, they will be predicted as \emph{inside}. Otherwise, the angle between the centers of the boxes is calculated to determine whether the prediction is \emph{above}, \emph{below}, or \emph{beside}.

\subsection{Image Synthesis}
\label{appendix-image}
We generate images of $512 \times 512$ pixels with text prompts. We use 1) VQGAN~\citep{esser2021taming}, which takes in a vector, and outputs a high-resolution image; and 2) CLIP~\citep{radford2021learning}, which can encode both text and images, and map them into a multi-modal embedding space. Image synthesis is the process of finding the optimal vector $\boldsymbol{v}$ inputted to VQGAN.
In each iteration, the vector is fed into VQGAN to generate an image $img=\textrm{VQGAN}(\boldsymbol{v})$. CLIP encodes the image into $\boldsymbol{c}=\textrm{CLIP}(img)$, and encodes the text prompt into $\boldsymbol{t}=\textrm{CLIP}(text)$, respectively. 

The optimization goal is to bring $\boldsymbol{c}$ and $\boldsymbol{t}$, the representation of the image and text encoded by CLIP closer.
The vector $\boldsymbol{v}$ is randomly initialized and optimized for 600 iterations. We use Adam optimizer with a learning rate of $0.5$. This process looks like a normal model ``training'', but here both VQGAN and CLIP are pretrained and their parameters are frozen; only the vector $\boldsymbol{v}$ is optimized from randomness for every prompt.

\subsection{Prompt Candidates Generation}
\label{appendix-prompt}
When probing PLMs, we follow~\citet{jiang2020can} to generate prompt and answer candidates with back-translation. Manually designed prompts and answers are translated from English to German and then backward. It is used to construct candidates with similar meanings. We leverage the translation model designed in~\citet{ng2019facebook}.

\subsection{Computing Infrastructure}
Experiments are conducted on NVIDIA GeForce RTX 3090 GPU. It takes 8 hours to generate 500 images on one GPU, and all other experiments can be executed in a few minutes.

\section{Probing Results on RelativeSize}
\label{appendix-relativesize}
\begin{table}[t]
    \centering
    \small
    \begin{tabular}{lcc}
    \toprule
    \textbf{Model} & \textbf{Acc} (avg. / $\sigma$) & \textbf{F1} (avg. / $\sigma$)\\
    \midrule
    BERT & 49.0 / 4.11 & 43.7 / 8.25 \\
    RoBERTa & 48.9 / 1.71 & 43.4 / 5.42\\
    VinVL & \textbf{60.6} / 1.47 & \textbf{51.2} / 2.22 \\
    \midrule
    \textbf{Model} & \textbf{Acc} & \textbf{F1}\\
    \midrule
    Best PLM & 49.0 (47.5) & 43.7 (40.5) \\
    VinVL & \textbf{60.6} (60.8) & 51.2 (49.8) \\
    ISM (Box) & 58.5 (\textbf{71.5}) & \textbf{58.5} (\textbf{71.4}) \\
    \midrule
    Best PLM & 49.0 (48.5) & 43.7 (43.5) \\
    VinVL & 60.6 (65.5) & 51.2 (55.7) \\
    ISM (Human) & \textbf{72.5} (\textbf{76.5}) & \textbf{71.8} (\textbf{75.7}) \\
    \bottomrule
    \end{tabular}
    \caption{Probing performance on RelatizeSize. Accuracy and macro F1 are reported. The numbers are in percentages (\%). In the last six lines, the first number is the performance on the whole dataset, and the number in parentheses indicates performance on the subset of instances where the generated images can be recognized by object detection models and humans, respectively. The standard deviation on different folds is represented with $\sigma$. Both objects are recognized with bounding boxes in 40\% images and are recognized by humans in 85\% images.}
    \label{table-relativesize}
\end{table}
RelativeSize~\citep{bagherinezhad2016elephants} is another dataset for comparing objects' sizes.
Table~\ref{table-relativesize} demonstrates the probing results on it. The results are consistent with those on our datasets: ISM probing, both evaluated with bounding boxes and evaluated by humans, outperforms PLM probing.

The methods used in~\citet{bagherinezhad2016elephants} are all retrieval-based. They execute search engine queries and download images from Flickr to make the comparisons. So we do not compare with their results directly. 
However, it is worth noticing that our ISM probing is comparable to the image retrieval-based baseline (its accuracy is 72.4\%). It exhibits that ISM learns sufficient knowledge from images.

\end{document}